\documentclass[letterpaper]{article} 
\usepackage[preprint]{aaai2027} 
\usepackage[hyphens]{url} 
\usepackage{graphicx} 
\usepackage{natbib} 
\usepackage{caption} 
\usepackage{booktabs}
\usepackage{amsmath}
\usepackage{amssymb}

\title{FocusGS: Spatial Delta Layers for Local Repair and Deterministic Editing of Trained 3D Gaussian Assets}
\author{Yiqun Pan, Yukun Shi\corresponding}
\affiliations{
College of Information Science and Technology, Beijing University of Chemical Technology, Beijing, China\\
\texttt{2024030483@buct.edu.cn}, \texttt{yukun\_shi@buct.edu.cn}
}

\begin{document}
\maketitle

\begin{abstract}
3D Gaussian Splatting (3DGS) is evolving from one-time reconstruction into deliverable, inspectable, and maintainable visual assets. Existing workflows focus on global reconstruction, training-time density control, or open-ended generative editing, leaving trained assets without precise local maintenance. We propose FocusGS, which unifies local repair and deterministic editing as composite spatial deltas. Repair is the purely additive special case: its base-manipulation term is empty, and it adds only local Gaussian bases; deterministic editing uses erase-insert factorization (EIF) to combine old-carrier erasure with new-content insertion. FocusGS addresses spatial gradient starvation: local repair raises target-region PSNR by 7.91 dB over 93 evaluation views. Across all 83 deterministic editing trials, the target ROI improves, with a trial-averaged mean edited ROI PSNR of 21.97 dB and a mean gain of +11.05 dB; across five public editing cases, FocusGS-EIF reaches 33.17 dB Target-mask PSNR and 0.994 Target-delta Correlation, while both text-driven baselines fail to complete the prescribed updates. FocusGS provides a lightweight, verifiable 3DGS maintenance operator.
\end{abstract}

\begin{figure}[!h]
  \centering
  \includegraphics[width=0.79\columnwidth]{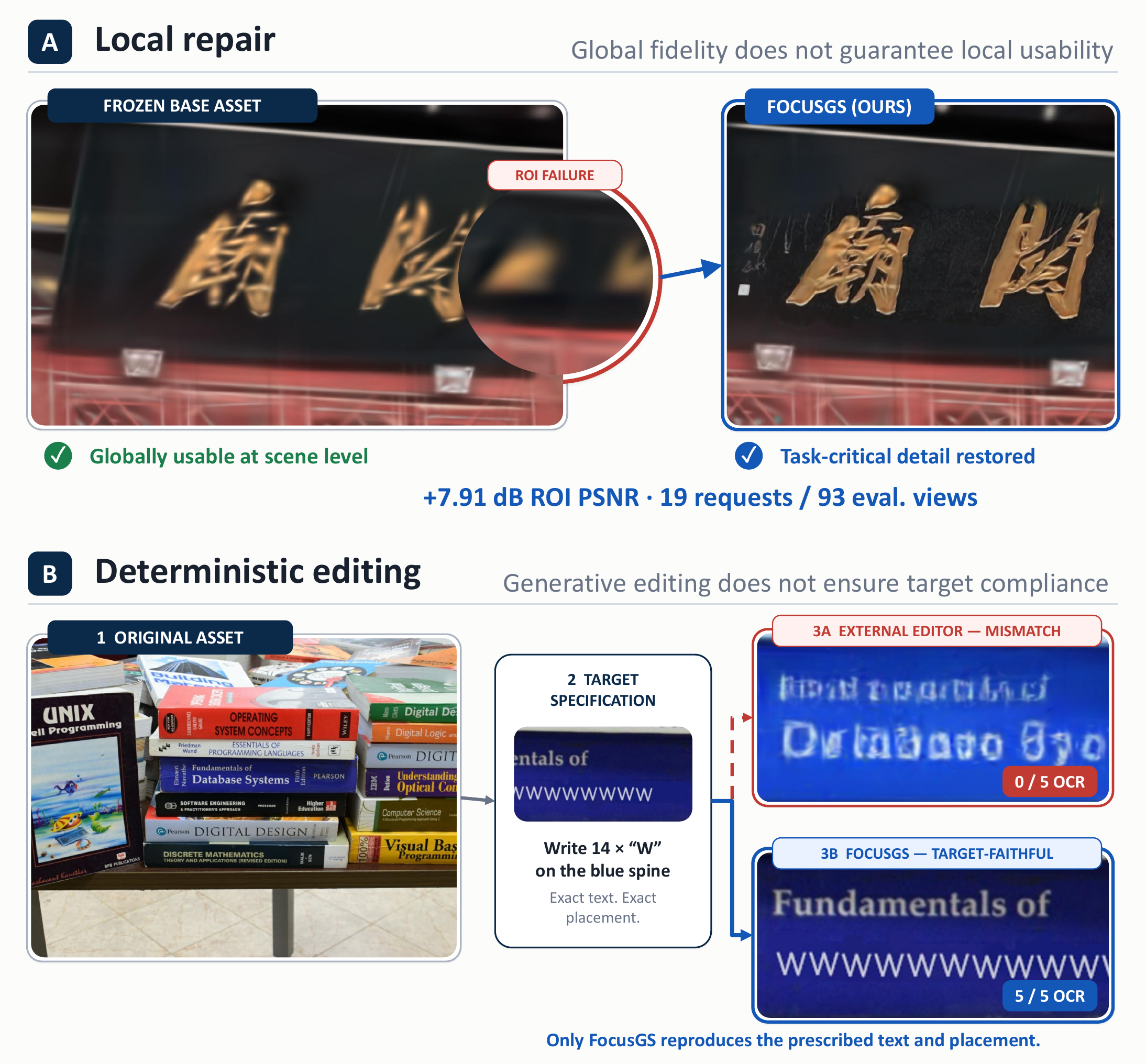}
  \captionsetup{skip=2pt}
  \caption{FocusGS restores observation-supported local detail (A) and performs prescribed target-faithful editing (B).}
  \label{fig:teaser}
\end{figure}

\newpage
\section{Introduction}
Large-scale reconstruction now involves thousands to tens of thousands of high-resolution images in aerial surveys, urban digital twins, and industrial capture. At this scale, 3DGS remains constrained by memory and long whole-scene optimization~\citep{lin2024vastgaussian,li2023matrixcity}. Expanding the global Gaussian budget or extending training may improve average fidelity, but it further increases whole-scene cost and still does not reserve optimization capacity for any particular local structure. As these scenes become reusable assets, they must be delivered, edited, and maintained.

This asset-oriented transition exposes a critical blind spot: global average quality is not equivalent to asset usability. A few high-information local regions can block delivery even when global PSNR or SSIM is strong. Sign text, product labels, architectural details, and fine textures occupy little image area but carry high task value; one unreadable detail can invalidate an otherwise strong asset.

Existing reconstruction methods mainly improve representation quality during global training, for example by adjusting density control, anti-aliasing, Gaussian placement, gradient aggregation, or global state transitions. They can improve average reconstruction metrics on their respective benchmarks, but they have not formed a local maintenance operator for trained assets. A deeper mismatch comes from the optimization object: global training optimizes the average error of the whole image, whereas a maintenance request after deployment requires concentrating on the repair of one small region under strong constraints. Reactivating the entire scene for one local defect reallocates the limited budget to already satisfactory regions and can further consolidate incorrect local support. We term the resulting attenuation of small-area, high-value signals under full-image optimization and global capacity competition \emph{spatial gradient starvation}.

The same mismatch also appears in 3DGS editing. Many requests in real asset maintenance are deterministic: replacing specified text, updating a fixed logo, or accurately writing an already designed graphic into an existing asset. GaussianEditor and DGE target text-driven open-ended editing; under the strict target re-rendering protocol in this paper, although they produce obvious visual changes, neither writes the specified characters and layout in any test case: both obtain 0/5 OCR hits, and Target-delta Correlation is close to 0. For deterministic asset maintenance, incorrect content means a failed update.

Deterministic editing also faces a problem specific to 3DGS. Gaussian rendering uses front-to-back alpha compositing; old text is jointly represented by spatial Gaussians that participate in opacity accumulation and depth sorting and cannot be directly covered like a two-dimensional texture. If the old-content carriers remain active, directly stacking newly added Gaussians may produce mixed characters, color contamination, or side-view residues. Therefore, FocusGS uses EIF: it first identifies and cleans the old-content carriers, obtains an intermediate base, and then inserts new content on that base.

Based on the observations above, this paper proposes FocusGS, a spatial-delta-layer method for trained 3DGS assets. FocusGS unifies two types of maintenance requests as task-conditioned composite spatial deltas: repair corresponds to purely additive local basis expansion; deterministic editing corresponds to the combined update of old-carrier erasure and new-content insertion. The two tasks share multi-view support, context protection, local optimization, and held-out-view verification. Gradients and optimizer states are built only for the local delta parameters required by the current request, while the global asset outside the target region remains unchanged. Figure~\ref{fig:teaser} gives an overview of the two local maintenance tasks considered in this paper.

The contributions of this paper are as follows:
\begin{enumerate}
\item We identify and formalize spatial gradient starvation: full-image averaging attenuates task-critical local gradients by the ROI area ratio, leaving small regions under-supported under a fixed global budget.
\item We propose composite spatial-delta modeling: each request is jointly described by a base-manipulation term and a Gaussian-addition term, and repair is the purely additive special case in which the former is empty, directly adding missing local basis functions to the target region. FocusGS achieves +7.91 dB ROI PSNR over 93 evaluation views, versus $-1.43$ dB for continued global training and +2.35 dB for local-unfreeze.
\item For deterministic text and graphic updates, we propose footprint-aware erase-insert factorization (EIF), explicitly decomposing old-content-carrier erasure and new-content insertion. Across 83 deterministic editing trials, FocusGS improves the target ROI in every run; across five public editing cases, it achieves 5/5 OCR hits and 0.994 Target-delta Correlation, whereas both text-driven external methods obtain 0/5.
\end{enumerate}

\section{Related Work}
\paragraph{3DGS reconstruction and training-time density control.}
The original 3DGS uses anisotropic Gaussians, differentiable rasterization, and alternating density control to achieve high-quality real-time rendering~\citep{kerbl2023_3dgs}. Subsequent methods improve global training and Gaussian configurations through anti-aliasing, pixel-aware gradients, homodirectional gradients, and MCMC state transitions~\citep{yu2024mipsplatting,zhang2024pixelgs,ye2024absgs,kheradmand2024mcmc}. These methods answer ``how to train the entire scene better,'' but do not provide a local maintenance operator for a small number of critical regions after training. Average gains can still leave text, signs, and fine textures unreadable. STRinGS studies selective text refinement~\citep{raundhal2026strings}; FocusGS unifies observation-supported repair and prescribed erase-insert editing under held-out verification.

\paragraph{Explicit Gaussian manipulation.}
Tools such as SuperSplat~\citep{playcanvas2026supersplat} and Splatshop~\citep{schutz2025splatshop} generally use an AABB, screen-space selections, or instance masks to extract an existing Gaussian set and then perform translation, deletion, scaling, duplication, or recoloring. Splatshop positions itself as a step toward ``Photoshop for Gaussian Splatting.'' Such tools are suitable for scene cleaning and object rearrangement, but they manipulate existing primitives; they neither learn missing local bases from target evidence nor write prescribed text, logos, or fine-grained textures with pixel-aligned fidelity.

\paragraph{Generative semantic editing.}
The GaussianEditor series and DGE share the basic route of ``2D generative editing--multi-view consistency--3D Gaussian write-back,'' with differences in region localization, cross-view propagation, and optimization strategy~\citep{chen2024gaussianeditor,wang2024gaussianeditor,chen2024dge}. Their objectives prioritize semantic plausibility and instruction-guided appearance changes rather than pixel-aligned reproduction of a prescribed target. Fixed characters, absolute placement, and strict layout therefore remain outside their objective. FocusGS supplies this missing deterministic maintenance operator through EIF.

\section{Method: Spatial Delta Layers}
\subsection{Problem Definition and Composite Spatial Delta}
Let the trained 3D Gaussian base asset be
\begin{equation}
G_0=\{g_i\}_{i=1}^{N}, \qquad
g_i=(\boldsymbol{\mu}_i,\boldsymbol{\Sigma}_i,\alpha_i,\mathbf{c}_i),
\label{eq:base}
\end{equation}
where $\boldsymbol{\mu}_i$, $\boldsymbol{\Sigma}_i$, $\alpha_i$, and $\mathbf{c}_i$ denote center, covariance, opacity, and color/spherical-harmonic coefficients. For view $v$, $R_v(\cdot)$ is the differentiable renderer. We assume $G_0$ is globally usable but contains a few regions requiring repair or deterministic updates.

Each local request is represented by a support bundle
\begin{equation}
\mathcal{T}=\{(v,M_v,Y_v)\}_{v\in\mathcal{V}_{\mathrm{sup}}},
\label{eq:bundle}
\end{equation}
where $\mathcal{V}_{\mathrm{sup}}$ provides geometry and visibility support, $M_v$ is the propagated soft mask, and $Y_v$ is the local target from real observations (repair) or a user-specified patch propagated through approximately planar geometry (editing).

\begin{figure*}[t]
  \centering
  \includegraphics[width=0.93\textwidth]{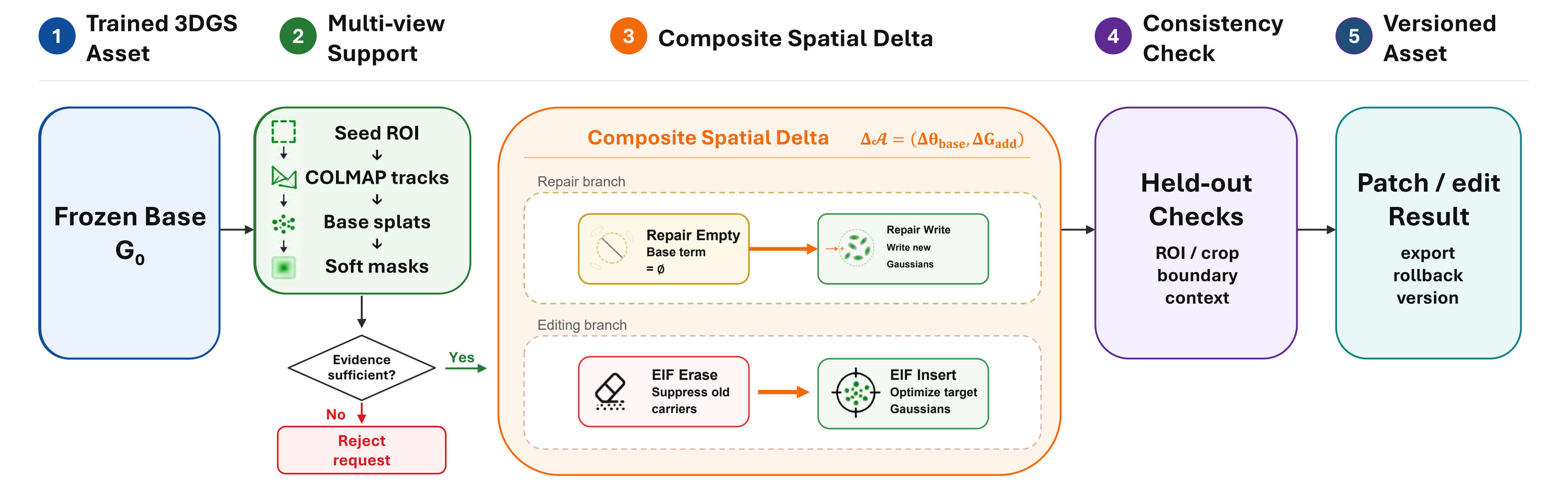}
  \caption{Overview of the FocusGS spatial-delta pipeline, including repair and deterministic editing branches.}
  \label{fig:pipeline}
\end{figure*}

As shown in Figure~\ref{fig:pipeline}, each request is a composite spatial delta with a base-manipulation term for locally suppressing, replacing, or refining selected Gaussians and a Gaussian-addition term for new local bases. Repair is the purely additive special case: the base remains frozen and only the repair delta is optimized:
\begin{equation}
G_{\mathrm{rep}}=G_0\oplus\Delta G_{\mathrm{rep}},
\qquad \nabla_{G_0}\mathcal{L}=0,
\label{eq:repairdelta}
\end{equation}
where $\Delta G_{\mathrm{rep}}$ is the added local Gaussian set, and $\oplus$ denotes concatenation of Gaussian sets rendered jointly. Editing uses the complete delta: $\Delta\theta_{\mathrm{erase}}$ produces a cleaned intermediate base, and $\Delta G_{\mathrm{ins}}$ inserts new content:
\begin{equation}
\begin{aligned}
G_{\mathrm{clean}}&=\mathcal{E}(G_0;M_{\mathrm{erase}},M_{\mathrm{under}},M_{\mathrm{protect}}),\\
G_{\mathrm{edit}}&=G_{\mathrm{clean}}\oplus\Delta G_{\mathrm{ins}}.
\end{aligned}
\label{eq:eif}
\end{equation}
The erasure delta acts on selected carriers through index suppression, appearance replacement, or optional local refinement; the cleaned base is then fixed while optimizing $\Delta G_{\mathrm{ins}}$. The unified representation is
\begin{equation}
\Delta\mathcal{A}_q=(\Delta\theta_{\mathrm{base}},\Delta G_{\mathrm{add}}),
\qquad
G^\star=\operatorname{Apply}(G_0,\Delta\mathcal{A}_q).
\label{eq:unified}
\end{equation}
Repair sets $\Delta\theta_{\mathrm{base}}=\varnothing$ and $\Delta G_{\mathrm{add}}=\Delta G_{\mathrm{rep}}$; editing sets $\Delta\theta_{\mathrm{base}}=\Delta\theta_{\mathrm{erase}}$ and $\Delta G_{\mathrm{add}}=\Delta G_{\mathrm{ins}}$. Both share multi-view support, local protection, and evaluation. The global base outside the target neighborhood is not optimized; the original checkpoint is retained, and repair residuals can be stored and unloaded independently.

\subsection{Mechanism Motivation}
\paragraph{Spatial Gradient Starvation.}
Standard 3DGS optimizes the average reconstruction loss over the complete image domain $\Omega_v$:
\begin{equation}
\mathcal{L}_{\mathrm{global}}
=\sum_v\frac{1}{|\Omega_v|}
\sum_{p\in\Omega_v}\ell\!\left(R_v(G)(p),I_v(p)\right).
\label{eq:global}
\end{equation}
\noindent\textbf{Proposition 1 (Area-ratio attenuation under local support).}
Suppose a parameter set $\theta_M$ mainly affects a small region $M_v\subset\Omega_v$. Under the locality assumption that gradients outside the region are approximately zero, the contribution of $M_v$ to the full-image gradient is scaled by $|M_v|/|\Omega_v|$:
\begin{equation}
\nabla_{\theta_M}\mathcal{L}_{\mathrm{global}}
=\sum_v\frac{|M_v|}{|\Omega_v|}
\left(
\frac{1}{|M_v|}\sum_{p\in M_v}\nabla_{\theta_M}\ell_{v,p}
\right).
\label{eq:attenuation}
\end{equation}
When $|M_v|\ll|\Omega_v|$, full-image averaging suppresses the local signal before capacity is redistributed. Under limited iterations, memory, and Gaussian capacity, small high-frequency residuals cannot dominate parameter and support redistribution. FocusGS freezes the usable global part and concentrates new capacity, gradients, and optimizer states on the local delta parameters that determine asset usability. Table~\ref{tab:alternatives} and Figure~\ref{fig:curves} expose the operational consequence of this attenuation.

\paragraph{Old-Content Leakage under Alpha Compositing.}
For view $v$, let $\mathcal{O}$ denote retained old-content carriers. Their contribution can be isolated as
\begin{equation}
\begin{aligned}
C_v(p)&=C_v^{\neg\mathcal{O}}(p)+r_v^{\mathcal{O}}(p),\\
r_v^{\mathcal{O}}(p)&=\sum_{i\in\mathcal{O}}T_{i,v}(p)\alpha_{i,v}(p)\mathbf{c}_i,\\
T_{i,v}(p)&=\prod_{j<i}\left(1-\alpha_{j,v}(p)\right).
\end{aligned}
\label{eq:leakage}
\end{equation}
Here, the first term sums non-carrier contributions under the original depth order and transmittances.
\noindent\textbf{Proposition 2 (Residual-carrier leakage).}
Active old carriers leave a view-dependent residual. Because footprint, transmittance, and depth order vary across views, fitting one view does not guarantee removal of residues in others; EIF suppresses the carriers first.

\subsection{Multi-View Local Support Bundle}
A single-view crop objective admits view-dependent billboard solutions. FocusGS builds a multi-view bundle from the seed mask using COLMAP points and tracks together with base-Gaussian projections; sparse SfM support falls back to visible base projections, followed by dilation and feathering. For approximately planar carriers such as books and signboards, an estimated homography propagates role masks and target patches.

Targets may come from real observations, manual retouching, interactive edits, or manually confirmed generated images. EIF assimilates the definite local target into a 3DGS update.

\begin{figure}[t]
  \centering
  \includegraphics[width=\columnwidth]{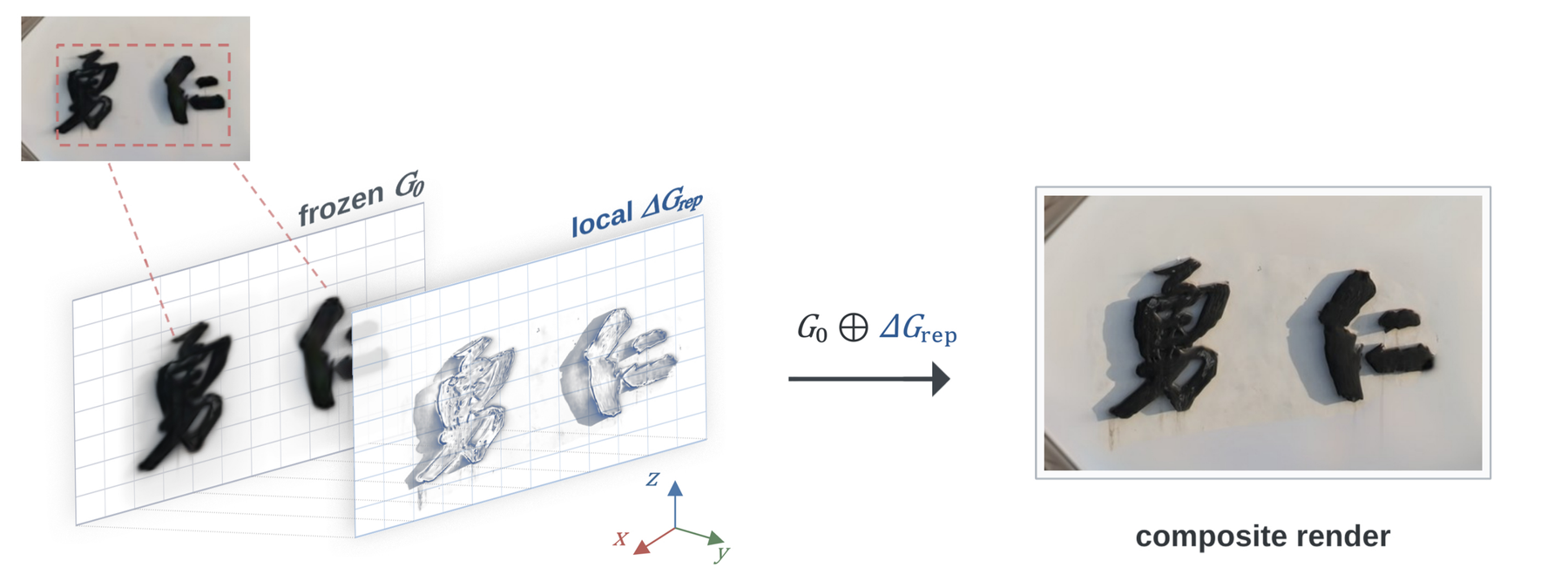}
  \caption{Spatial-delta realization for local repair. The trained base $G_0$ remains frozen, while a compact local repair layer $\Delta G_{\mathrm{rep}}$ supplies the missing structures; their composition produces the repaired render.}
  \label{fig:repairlayer}
\end{figure}

\subsection{Local Repair Delta: Adding Missing Representation Bases}
The repair delta layer consists of a set of newly added Gaussians initialized inside or near the target support region. The base Gaussians are detached from the computational graph, and the optimizer updates only the positions, scales, rotations, opacities, and appearance of $\Delta G_{\mathrm{rep}}$. The objective is
\begin{equation}
\mathcal{L}_{\mathrm{rep}}
=\mathcal{L}_{\mathrm{target}}
+\lambda_p\mathcal{L}_{\mathrm{protect}}
+\lambda_o\mathcal{L}_{\mathrm{opacity}}
+\lambda_s\mathcal{L}_{\mathrm{scale}},
\label{eq:repairloss}
\end{equation}
where
\begin{equation}
\mathcal{L}_{\mathrm{target}}
=\frac{1}{|\mathcal{V}_{\mathrm{sup}}|}
\sum_{v\in\mathcal{V}_{\mathrm{sup}}}
\frac{1}{|M_v|}
\sum_{p\in M_v}
[\ell_1(p)+\lambda_{\mathrm{ssim}}\ell_{\mathrm{ssim}}(p)].
\label{eq:targetloss}
\end{equation}
The protection loss prevents added Gaussians from altering context outside the ROI, while opacity and scale regularization suppress oversized or overly dense residuals. By expanding the local function space beyond fixed existing bases, the repair delta represents missing strokes, edges, and fine textures that local unfreezing cannot supply. Projected footprints are used only for EIF carrier selection, not as a repair-loss regularizer. Figure~\ref{fig:repairlayer} visualizes this frozen-base/local-delta composition.

\subsection{EIF: Footprint-Aware Erase-Insert Factorization}
The input contains the original, clean-background, and modified target images. Image differences define functional masks
$M_{\mathrm{erase}}=\mathbf{1}[\|I_{\mathrm{orig}}-I_{\mathrm{cover}}\|_\infty>\tau_e]$
and
$M_{\mathrm{insert}}=\mathbf{1}[\|I_{\mathrm{mod}}-I_{\mathrm{cover}}\|_\infty>\tau_i]$;
bounding boxes, dilation, and neighborhood differences construct $M_{\mathrm{under}}$ and $M_{\mathrm{protect}}$.

For base Gaussian $g_i$, let $B_{i,v}$ be the lightweight 2D footprint from its projected center and radius. We define
\begin{equation}
\begin{aligned}
S_{i,v}(M)&=\operatorname{Reduce}_{p\in B_{i,v}}M(p),\\
\rho_{i,v}&=
\frac{S_{i,v}(M_{\mathrm{erase}})}
{S_{i,v}(M_{\mathrm{erase}})+S_{i,v}(M_{\mathrm{protect}})+\varepsilon}.
\end{aligned}
\label{eq:footprint}
\end{equation}
where Reduce is center, maximum-footprint, or mean-footprint sampling. Old carriers are selected by footprint, protection ratio, visibility, opacity, projected radius, and optional depth consistency. The lightweight 2D score approximates projected participation without full pixel-level volumetric attribution. Stage 1 suppresses selected indices, replaces their appearance with the clean background, or optionally refines them; Stage 2 freezes the cleaned base and trains the insertion layer with repair-style target and protection losses. This decomposition cuts off old-content contributions in alpha compositing and sharply reduces mixed characters, color contamination, and side-view residues. Direct insertion without erasure, repair-as-editing, and replacement of the corresponding COLMAP training images all produced severe old-content residue or unstable cross-view propagation, supporting the erase-insert design.

\subsection{Verifiability and Versioned Delivery}
The implementation holds out evaluation views and verifies that they never enter training. Repair reports ROI, crop, non-target, and boundary-ring metrics; editing additionally reports Target-mask PSNR, Target Edge-F1, Target-delta Correlation, OCR hits, and old-content residues. This multi-view regression check evaluates whether improvements extend beyond the seed view while preserving context in unseen views.

FocusGS rejects a request before optimization when the seed view is unregistered or reliable multi-view support cannot be recovered. In batch mode, segmentation can generate candidate masks, multi-view tracking deduplicates them, and FocusGS processes the remaining requests. Repair residuals are independently stored; EIF preserves the original checkpoint and outputs the cleaned base and insertion result for version-level rollback.

\section{Experiments}
\subsection{Setup and Evaluation Protocol}
\paragraph{Local repair.}
We evaluate FocusGS on a high-resolution DJI asset built from sharp observations: over 4,000 GPS-tagged $5280\times3956$ images covering about $110\ \mathrm{m}\times150\ \mathrm{m}$, with MCMC-2M as the strong base. Each of 23 user-specified requests is screened for camera/SfM support. FocusGS optimizes 19 supported requests for 3,000 steps and rejects four before optimization when the seed view is unregistered or reliable multi-view support cannot be recovered. Results first average views within each request, then average the 19 request means; improvement counts cover all 93 evaluation views.

\paragraph{Repair metrics.}
ROI/Crop PSNR measure target fidelity and local context; Edge-F1, GradCorr, StrokePSNR, and StrokeMAE measure stroke and gradient structure; Non-ROI and boundary-ring changes measure context safety.

\paragraph{Deterministic editing.}
For each public case, all methods use the same target view, binary target mask, and resolution. Target-mask PSNR measures masked RGB error; Target Edge-F1 uses luminance Sobel edges with a masked 75th-percentile threshold (minimum 0.03) and exact-pixel F1; Target-delta Correlation compares masked output and target luminance residuals relative to the clean carrier; OCR requires complete recognition. The broader evaluation contains 83 prescribed-target trials and 408 evaluation views. A trial is one local target update; views are averaged within each trial, then the 83 trial means are averaged equally. The five public cases are reported separately for external comparison.

\begin{figure*}[t]
  \centering
  \includegraphics[width=0.94\textwidth]{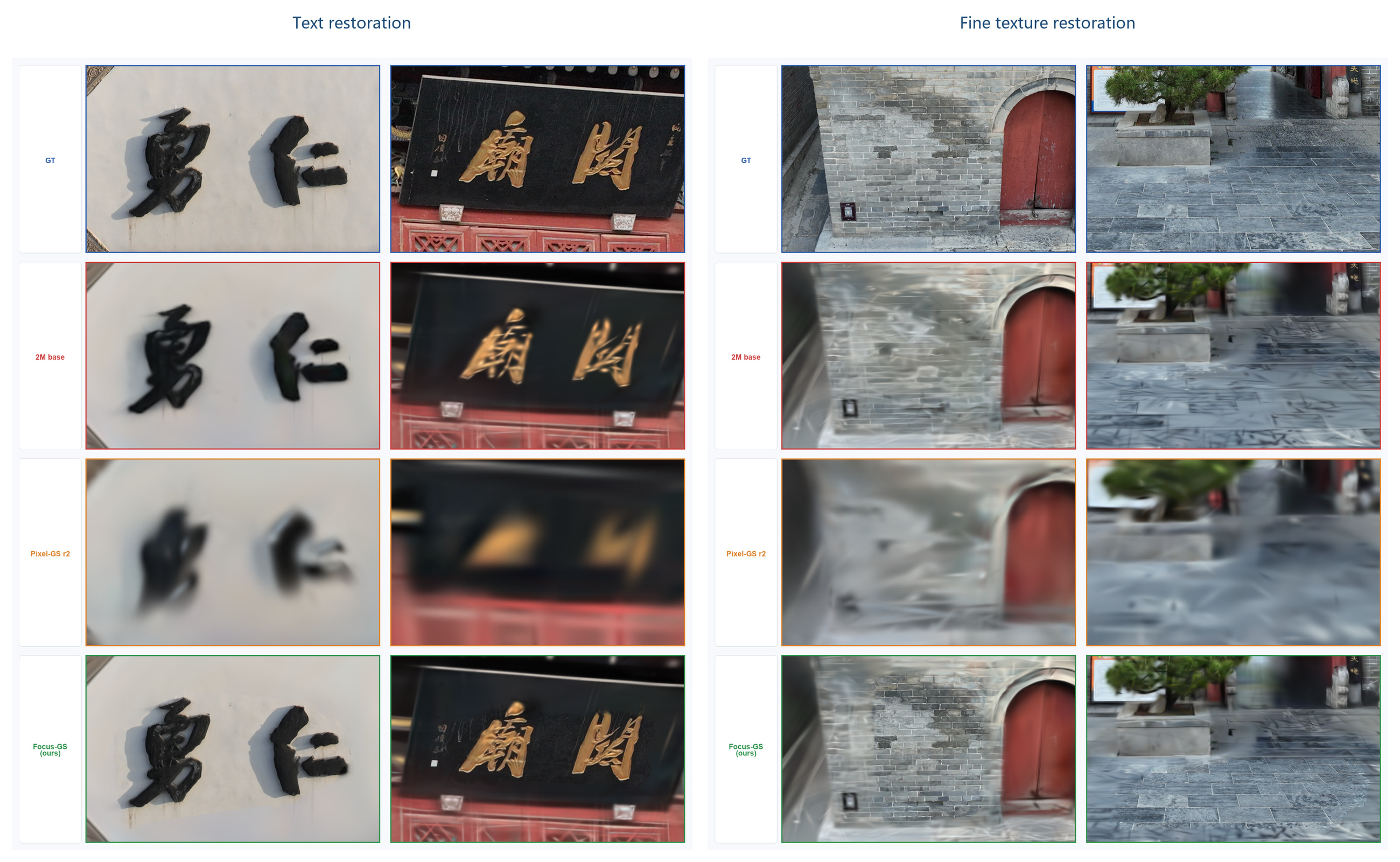}
  \caption{Local-repair comparison: GT, MCMC-2M, Pixel-GS, and FocusGS.}
  \label{fig:repaircomp}
\end{figure*}

\subsection{Main Local Repair Results and Post-Training Comparison}
Table~\ref{tab:repair} summarizes the main protocol. ROI PSNR and Edge-F1 improve in 93/93 views, Crop PSNR and StrokePSNR in 90/93, GradCorr in 91/93, and StrokeMAE in 92/93. Mean ROI and Crop PSNR gains reach 7.91 and 3.12 dB; the left text case in Figure~\ref{fig:repaircomp} reaches 13.63 and 8.54 dB. FocusGS restores observed stroke topology and fine texture in held-out views.

\begin{table}[t]
\centering
\small
\setlength{\tabcolsep}{3.2pt}
\begin{tabular}{llll}
\toprule
Metric & Evidence & Mean $\Delta$ & Improved\\
\midrule
ROI PSNR & target fidelity & +7.913 dB & 93/93\\
Crop PSNR & local context & +3.123 dB & 90/93\\
Edge-F1 & edge topology & +0.437 & 93/93\\
GradCorr & gradient structure & +0.302 & 91/93\\
StrokePSNR & stroke fidelity & +3.11 dB & 90/93\\
StrokeMAE $\downarrow$ & stroke error & -0.039 & 92/93\\
\bottomrule
\end{tabular}
\caption{Local repair results on 19 completed requests (93 evaluation views; 23/19/4 total/completed/rejected).}
\label{tab:repair}
\end{table}

\begin{table}[t]
\centering
\small
\setlength{\tabcolsep}{3.0pt}
\begin{tabular}{lrrrr}
\toprule
Method & Update & $\Delta$ROI & $\Delta$Crop & Time\\
\midrule
MCMC-2M & -- & -- & -- & --\\
Continue & whole base & -1.430 & -1.496 & 59.77 s\\
Local-unfreeze & selected base & +2.351 & +1.375 & 27.15 min\\
\textbf{FocusGS} & \textbf{new bases} & \textbf{+7.913} & \textbf{+3.123} & \textbf{4.63 min}\\
\bottomrule
\end{tabular}
\caption{Post-training alternatives on 19 requests and 93 views.}
\label{tab:alternatives}
\end{table}

Table~\ref{tab:alternatives} separates three post-training routes. Continue resumes whole-scene optimization; Local-unfreeze updates existing Gaussians selected by the propagated support, without Gaussian birth or densification; FocusGS adds new local bases. Continue decreases ROI/Crop PSNR by 1.430/1.496 dB. Local-unfreeze recovers part of the signal (+2.351/+1.375 dB) but remains constrained by fixed support. FocusGS reaches +7.913/+3.123 dB, isolating explicit basis expansion as the source of the larger recovery. Crop non-ROI remains stable (+0.17 dB), and the boundary ring improves by +2.99 dB.

\begin{figure*}[t]
  \centering
  \includegraphics[width=0.65\textwidth]{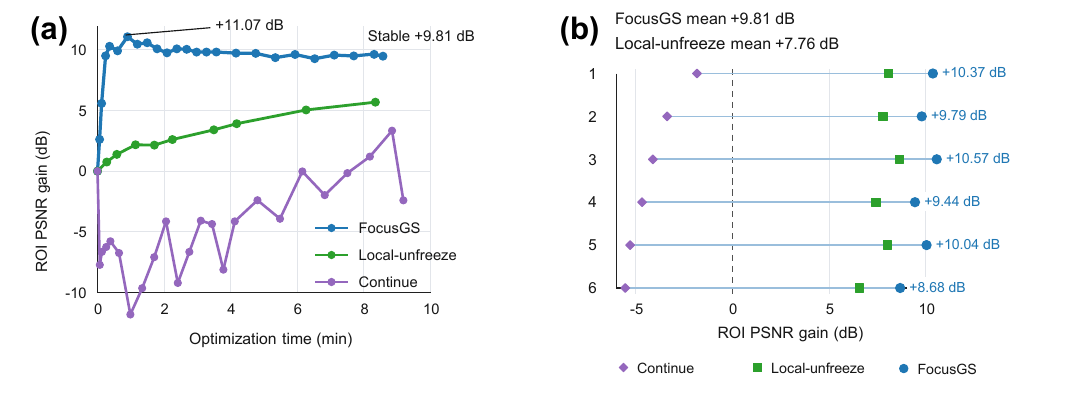}
  \caption{Early convergence and final gains on PS17. (a) First 10 min. (b) Final six-view gains.}
  \label{fig:curves}
\end{figure*}

On the same 19-case protocol, mean training time is 27.15 min for Local-unfreeze and 4.63 min for FocusGS. Figure~\ref{fig:curves}(a) shows the six-view PS17 diagnostic: FocusGS reaches a stable +9.81 dB at 3,000 steps; panel (b) reports final per-view gains. Together, the aggregate and trajectory results show that FocusGS reaches the stronger solution with substantially less optimization time. These results establish a practical post-training maintenance regime.

\subsection{Deterministic Editing and Target Compliance}
Existing public 3DGS editors optimize open-ended, text-guided changes rather than prescribed target reproduction. We test GaussianEditor~\citep{wang2024gaussianeditor} and DGE~\citep{chen2024dge} as deterministic maintenance operators on five cases from four public datasets: Books-I and Books-II, two book-spine targets from the same STRinGS-360 Books scene~\citep{raundhal2026strings}, Train from Tanks and Temples~\citep{knapitsch2017tanks}, Playroom from Deep Blending~\citep{hedman2018deep}, and one vehicle case from 3DRealCar~\citep{du2025realcar}. Books-III is a third target from the same Books scene used only for EIF ablation. For 3DRealCar, we reconstruct the 3DGS asset from its released multi-view images. Both baselines run simple, standard, and complex prompts; each case fixes one PSNR-best output, on which all metrics are computed.

Across 83 trials (408 views), all 83 trials show a positive mean target-ROI gain. Trial-averaged edited ROI PSNR is 21.97 dB, with a +11.05 dB gain. Across the five public cases, FocusGS-EIF reaches 33.17 dB Target-mask PSNR, 0.967 Edge-F1, 0.994 Target-delta Correlation, and 5/5 OCR hits. The two baselines remain near 10 dB, about 0.26 Edge-F1, near-zero correlation, and 0/5 OCR.

\begin{figure*}[t]
  \centering
  \includegraphics[width=0.89\textwidth]{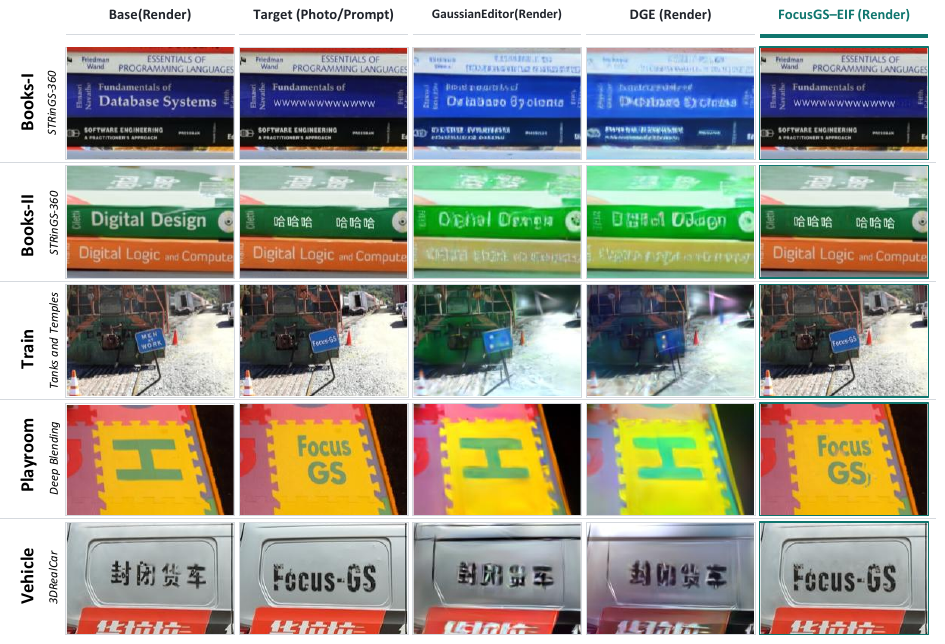}
  \caption{Deterministic-editing comparison on five public editing cases.}
  \label{fig:editing}
\end{figure*}

Figure~\ref{fig:editing} shows the qualitative comparison, while Table~\ref{tab:editing} reports the corresponding per-case metrics.

\begin{table}[!h]
\centering
\small
\setlength{\tabcolsep}{3.7pt}
\begin{tabular}{llrrrr}
\toprule
Case & Method & PSNR & E-F1 & Corr. & OCR\\
\midrule
Books-I & GE & 11.89 & 0.269 & -0.037 & 0\\
 & DGE & 11.62 & 0.276 & -0.038 & 0\\
 & \textbf{Ours} & \textbf{35.31} & \textbf{0.973} & \textbf{0.998} & \textbf{1}\\
\midrule
Books-II & GE & 8.92 & 0.266 & 0.024 & 0\\
 & DGE & 8.83 & 0.263 & 0.047 & 0\\
 & \textbf{Ours} & \textbf{30.31} & \textbf{0.965} & \textbf{0.993} & \textbf{1}\\
\midrule
Train & GE & 10.27 & 0.315 & -0.019 & 0\\
 & DGE & 9.42 & 0.243 & -0.050 & 0\\
 & \textbf{Ours} & \textbf{33.31} & \textbf{0.964} & \textbf{0.993} & \textbf{1}\\
\midrule
Playroom & GE & 9.84 & 0.245 & 0.040 & 0\\
 & DGE & 9.72 & 0.264 & 0.052 & 0\\
 & \textbf{Ours} & \textbf{32.68} & \textbf{0.966} & \textbf{0.990} & \textbf{1}\\
\midrule
3DRealCar & GE & 9.79 & 0.251 & 0.034 & 0\\
 & DGE & 9.99 & 0.252 & -0.008 & 0\\
 & \textbf{Ours} & \textbf{34.24} & \textbf{0.967} & \textbf{0.997} & \textbf{1}\\
\bottomrule
\end{tabular}
\caption{Per-case results. GE is GaussianEditor~\citep{wang2024gaussianeditor}; E-F1 denotes Target Edge-F1.}
\label{tab:editing}
\end{table}

\paragraph{Mechanism ablation.}
We isolate erasure on Books-I and Books-III, two book-spine targets from the same STRinGS-360 Books scene. Direct insertion and complete EIF use identical targets, masks, views, optimization settings, and seeds; only the input checkpoint differs. Complete EIF reduces the old-region L1 residue from 0.099 to 0.008 and raises Crop PSNR from 26.71 to 31.66 dB, while preserving comparable Target-mask PSNR (28.33 versus 28.48 dB). This paired control confirms that erasure removes old carriers without sacrificing target insertion.

\begin{table}[!h]
\centering
\small
\setlength{\tabcolsep}{4.0pt}
\begin{tabular}{llrrr}
\toprule
Case & Variant & T-mask & Old L1 $\downarrow$ & Crop $\uparrow$\\
\midrule
Books-I & Direct & 29.23 & 0.101 & 28.23\\
 & Complete EIF & 28.86 & \textbf{0.010} & \textbf{31.92}\\
Books-III & Direct & 27.72 & 0.098 & 25.20\\
 & Complete EIF & 27.79 & \textbf{0.006} & \textbf{31.39}\\
\bottomrule
\end{tabular}
\caption{EIF ablation on Books-I/III from the same STRinGS-360 Books scene; T-mask and Crop are PSNR (dB).}
\label{tab:ablation}
\end{table}

The near-zero target correlation and 0/5 OCR compliance show that open-ended generative editors are not deterministic maintenance operators. FocusGS-EIF is the only compared method that satisfies fixed character, position, and layout constraints across all five public editing cases.

\subsection{Efficiency and Operability on Large-Scale Assets}
FocusGS converts scene-wide retraining into lightweight local updates. The asset contains over 4,000 roughly 20-MP aerial images. Under our settings, a 2M base trains on an RTX 5070 under tight memory, while full 8M training requires an A100. By omitting base gradients and optimizer states, the same RTX 5070 maintains both bases in minutes. Independent deltas are about 10 MB, enabling versioned delivery and rollback.

On RTX 5070/8M, average wall/training time is 6.31/4.29 min and incremental peak memory is 4.13 GB. Across 2M--8M bases and both GPUs, maintenance remains minute-level; base size mainly affects loading and rasterization. Memory is baseline-subtracted from 1 Hz \texttt{nvidia-smi} traces; wall time includes support construction, evaluation, and file writing (Table~\ref{tab:runtime}).

\begin{table}[t]
\centering
\small
\setlength{\tabcolsep}{3.2pt}
\begin{tabular}{lrrrr}
\toprule
Device / Base & Wall & Train & Mean $\Delta$VRAM & Peak $\Delta$VRAM\\
\midrule
RTX 5070 / 2M & 4.87 & 2.80 & 2.27 & 2.60\\
RTX 5070 / 8M & 6.31 & 4.29 & 3.67 & 4.13\\
A100 / 2M & 5.85 & 2.75 & 3.10 & 3.87\\
A100 / 8M & 6.36 & 3.25 & 4.61 & 5.37\\
\bottomrule
\end{tabular}
\caption{Runtime (min) and incremental VRAM (GB) over three requests.}
\label{tab:runtime}
\end{table}

\FloatBarrier
\section{Applicability Boundaries and Failure Analysis}
FocusGS targets globally usable 3DGS assets whose local text, signs, labels, textures, or structures affect delivery and whose target regions provide recoverable evidence, clear carriers, and deterministic supervision. This boundary keeps updates verifiable and prevents generative guesses from being presented as reconstruction.

First, the method requires reliable observations, cameras, geometry, and visibility. In the motion-blurred, view-sparse STRinGS-360 books scene, Base 2M, Pixel-GS, AbsGS, and FocusGS patch obtain 10.23, 11.46, 11.46, and 10.25 dB, showing that extra local parameters cannot generate real details. Incorrect cameras, severe misalignment, or failed global reconstruction likewise remove the support required for a valid local update; in these cases, FocusGS rejects the request.

Second, the method best fits clear physical carriers such as wall text, book spines, signs, labels, cultural-relic details, tiles, and window grids with approximately planar or weakly curved texture. Multiple depths, transparency, strong reflection, non-rigidity, or severe occlusion may cause leakage, stretching, floating patches, or residues and require depth layering, finer geometry, or explicit surfaces.

Third, editing requires targets that are clear and alignable in character, position, layout, and local perspective. Generated targets require manual confirmation; otherwise, shape drift, edge misalignment, perspective changes, and texture rearrangement are directly assimilated into the asset.

\section{Conclusion}
Post-training local adaptation is a core maintenance capability missing from trained 3D Gaussian assets. FocusGS packages repair as additive local bases and EIF editing as old-carrier erasure plus target insertion. It restores text and fine textures across 93 views, improves the target ROI in all 83 deterministic editing trials, achieves 5/5 target compliance across five public editing cases while both text-driven baselines fail to complete the prescribed updates, and maintains 2M--8M assets with minute-level consumer-GPU overhead. FocusGS turns full-scene retraining into localized, verifiable, and deployable asset maintenance.

\clearpage
\bibliography{references}
\end{document}